\documentclass[twoside,twocolumn]{article}

\usepackage{graphicx}		 
\usepackage{hyperref}
\usepackage{amsmath}
\usepackage{amssymb}
\usepackage{amsfonts}
\usepackage[]{algorithm2e}
\usepackage[section]{placeins}
\usepackage{titling}
\graphicspath{ {images/} }
\usepackage[utf8]{inputenc}
\usepackage{breqn}
\usepackage{todonotes}

\newcommand*\samethanks[1][\value{footnote}]{\footnotemark[#1]}

\title{Adaptive Precision Training (AdaPT): 
\\A dynamic
quantized
training approach for DNNs}
\author{Lorenz Kummer\thanks{Faculty of Computer Science, University of Vienna} \thanks{lorenz.kummer@univie.ac.at} \and
Kevin Sidak \samethanks[1] \thanks{kevin.sidak@univie.ac.at}    \and
Tabea Reichmann \samethanks[1] \thanks{tabea.reichmann@univie.ac.at} \and
Wilfried Gansterer \samethanks[1] \thanks{wilfried.gansterer@univie.ac.at}}
\thanksmarkseries{arabic}
\date{August 2021}

\begin{document}
\maketitle
\begin{abstract}
Quantization is a technique for reducing deep neural networks (DNNs) training and inference times, which is crucial for training in resource constrained environments or applications where inference is time critical. State-of-the-art (SOTA) quantization approaches focus on \emph{post-training\/} quantization, i.e., quantization of pre-trained DNNs for speeding up inference. While work on quantized training exists, most approaches require refinement in full precision (usually single precision) in the final training phase or enforce a global word length across the entire DNN. This leads to suboptimal assignments of bit-widths to layers and, consequently, suboptimal resource usage. In an attempt to overcome such limitations, we introduce AdaPT, a new fixed-point quantized sparsifying training strategy. AdaPT decides about precision switches between training epochs based on information theoretic conditions. The goal is to determine on a \emph{per-layer basis\/} the lowest precision that causes no quantization-induced information loss while keeping the precision high enough such that future learning steps do not suffer from vanishing gradients. The benefits of the resulting fully quantized DNN are evaluated based on an analytical performance model which we develop. We illustrate that an average speedup of $1.27$ compared to standard training in float32 with an average accuracy increase of $0.98\%$ can be achieved for AlexNet/ResNet on CIFAR10/100 and we further demonstrate these AdaPT trained models achieve an average inference speedup of $2.33$ with a model size reduction of $0.52$.

\label{abstract}
\end{abstract}
\section{Introduction}
\label{sec:intro}
%
    
With the general trend in machine learning towards large model sizes to solve increasingly complex problems, inference in time critical applications or training under resource and/or productivity constraints is becoming more and more challenging. Applications, where time- and space efficient models are crucial, include robotics, augmented reality, self driving vehicles, mobile applications, applications running on consumer hardware, or scientific research where a high number of trained models is often needed for hyper parameter optimization. Accompanied by this, already some of the most common DNN architectures, like AlexNet~\cite{alexnet2012} or ResNet~\cite{he2016deep}, suffer from over-parameterization and overfitting~\cite{overparam2019allen, overparam2020li}

Possible solutions to the aforementioned problems include pruning (see, for example,~\cite{morphnet2018, squeezenet2016iandola, han2015deep, golub2018full, stewart2021nekmod}), or quantization. While network pruning, which aims at reducing the number of parameters in a given DNN architecture (e.g. by sparsification of weights tensors and using a sparse tensor format), is a successful strategy for reducing network size and runtime, it generally does not attempt to tailor parameter bit-width to exploit advantages available in low bit-width arithmetic, thus neglecting this potential to reduce computational resource consumption. We use pruning in the form of sparsification but our focus lies on making better use of quantization. When quantizing, the precision of the parameters and of the computation is decreased and the resultingmore coarse bit-width allows for more efficient use of computing resources and for runtime reductions.
However, quantization has to be performed with caution, as naive approaches or a too low bit-width can have a negative impact on the accuracy of the network and its ability to converge during training, which is unacceptable for important use cases. For example, binary quantization as proposed in~\cite{bnn2016} reduces memory requirements and can effectively speed up computation, as multiplication can be performed as bit shifts, but the accuracy suffers slightly from this approach and convergence is shown to much slower compared to float32 training.

Existing quantization approaches do not fully leverage the potential of quantized training. They do not take the differences quantization can have on different layers during training into account, nor dynamically raise or lower the precision used.
AdaPT extends these approaches and introduces a new precision switching mechanism based on the Kullback-Leibler-Divergence (KLD)~\cite{kullback1951information}, that calculates the average number of bits lost due to a precision switch. This is done not only for the entire network, but on a per-layer basis, therefore considering different quantization effects on the different layers, leading to \textbf{Ada}ptiv\textbf{E} \textbf{P}recision \textbf{T}raining of DNNs over time during training. Our approach does not need any refinement phase in full precision and produces an already fixed-point quantized network that can then also be deployed on high-performance application-specific integrated circuits (ASICs) or or field-programmable gate arrays (FPGAs) hardware.

\subsection{Related Work}
\label{ssec:related}
In general studies of the sensitivity of neural networks (NNs), perturbation analysis
has historically been applied with a focus on the theoretical sensitivity of single neurons or multi-layer perceptrons ~\cite{wangperturbation, meyerbaeseperturbation, zengperturbation}, but recently also yielded interesting results for the sensitivity of simple modern architectures by providing efficient algorithms for evaluating networks comparable to the complexity of LeNet-5~\cite{lenetperturbation, lenet1998}. While these studies certainly improve our understanding of the effect perturbation has on NNs, they lack practical applicability for quantized DNN training due to beeing largely analytical in nature.

For exploring the accuracy degradation induced by quantizations of weights, activation functions and gradients,~\cite{loroch2017tensorquant} and~\cite{zhang2019qpytorch} introduced the frameworks \emph{TensorQuant\/} and \emph{QPyTorch\/}, capable of simulating the most common quantizations for training and inference tasks on a float32 basis. Both frameworks allow to freely choose exponent and mantissa for floating-point representation, word and fractional bit length for fixed-point representation and word length for block-floating-point representation as well as signed/unsigned representations. Since quantization is only simulated in these frameworks, no runtime speedup can be achieved on this basis.

Several approaches have tried to minimize inference accuracy degradation induced by quantizing weights and/or activation functions while leveraging associated performance increases.
\emph{Quantization Aware Training\/} (QAT)~\cite{jacob2018QAT, yang2019QAT} incorporates simulated quantization into model training and trains the model itself to compensate the introduced errors. A similar approach is take by \emph{Uniform Noise Injection for Non-Uniform Quantization} (UNIQ)~\cite{baskin2021uniq}, which emulates a non-uniform quantizer to inject noise at training time to obtain a model suitable for quantized inference. \emph{Learned Quantization Nets\/} (LQ-Nets)~\cite{zhang2018ECCV} learn optimal quantization schemes through jointly training DNNs and associated quantizers. The \emph{Reinforcement-Learning Framework\/} (ReLeQ) introduced by~\cite{elthakeb2020releq} uses a reinforcement learning agent to learn the final classification accuracy w.r.t. the bit-width of each of the DNNs layers to find optimal layer to bit-width assignments for inference.
Dedicated quantization friendly operations are used in~\cite{quantconv2018}.
A different approach is taken by \emph{Variational Network Quantization\/}
(VNQ)~\cite{vnq2018achterhold} which uses variational dropout training~\cite{kingma2015variational} with a structured sparsity inducing prior~\cite{neklyudov2017structured} to formulate post-training quantization as the variational inference problem searching the posterior optimizing the KLD. 
\emph{High-order entropy minimization for neural network compression} (HEMP)~\cite{tartaglione2021hemp} introduces a entropy coding-based regularizer minimizing the quantized parameters entropy in gradient based learning and pairs it with a separate pruning scheme \cite{tartaglione2020lobster} to reach a high degree of model compression after training.

Machine learning frameworks such as PyTorch or Tensorflow already provide built-in quantization capabilities. These quantization methods focus on either QAT or on post-training quantization~\cite{tensorflow2015-quantization, pytorch-quantization}. Both of these methods only quantize the model after the training and consequently provide no efficiency gain during training. Additional processing steps during or after training are needed which add computational overhead.

From a theoretical perspective, quantized training has been investigated in~\cite{li2017training} with a particular focus on rounding methods and on convergence guarantees while~\cite{yin2019understanding} provided an analysis of the behaviour of the straight-through estimator (STE) ~\cite{bengio2013ste} during quantized training. For distributed training on multi-node environments, \emph{Quantized Stochastic Gradient Descent\/} (QSGD)~\cite{alistarh2017qsgd} incorporates a family of gradient compression schemes aimed at reducing inter-node communication occurring during SGD's gradient updates. This produces a significant reduction in runtime and network training can be guaranteed to converge under standard assumptions~\cite{alistarh2017qsgd_lg}. Very low bit width single-node training (binary, ternary or similarly quantized weights and/or activations), usually in combination with STE, has been shown to yield non-trivial reductions in runtime and model size as well but often either at comparably high costs to model accuracy, decreased convergence speed or not fully quantizing the network~\cite{bnn2016, hubara2017quantized, li2016ternary, yin2016quantization, hou2018lossaware, chu2021mixed} and its most capable representative, \emph{Quantized tensor train neural networks} (QTTNet)~\cite{lee2021qttnet}, which combines tensor decomposition and full low bit-width quantization during training to reduce runtime and memory requirements significantly, still suffers up to $2.5\%$ accuracy degradation compared to it's respective baseline. A low bit-width integer quantized (re-)~training approach reporting accuracy drops of $1\%$ or less compared to a float32 baseline was introduced by \cite{peng2021fullinteger}, but it requires networks pre-trained in full precision as starting points and it's explicitly stated goal is reducing computational cost and model size during inference. For speeding up training on a single node via block-floating point quantization,~\cite{muppet2020} introduced a dynamic training quantization scheme (\emph{Multi Precision Policy Enforced Training\/}, MuPPET). After quantized training it produces a model for float32 inference. 

\subsection{Contributions}
\label{ssec:motivation}
Existing quantized training and inference solutions leave room for improvements in several areas. QAT, LQ-Nets, ReLeQ, UNIQ, HEMP and VNQ are capable of producing networks such that inference can be performed under quantization with small accuracy degradation (relative to baselines, which are not comparable and where it is unclear how well they are optimized), but require computationally expensive float32 training and the algorithms themselves incur a certain overhead as well. QSGD training only quantizes gradients and focuses on reducing communication overhead in multi-node environments. Binary, ternary or similarly quantized training has been shown to commonly come at the cost of reduced accuracy, reduced convergence speed or only quantizing weights and not activations and its most capable representative QTTNet still does not achieve iso-accuracy. 

MuPPET requires at least $N$ epochs for $N$ quantization levels (pre-defined bit-widths applied globally to all layers) because precision switches are only triggered at the end of an epoch and the algorithm needs to go through all precision levels. Potential advantages from having different precision levels at different layers of the network are not exploited. The precision switching criterion is only based on the diversity of the gradients of the last $k$ epochs, no metric is used to measure the amount of information lost by applying a certain quantization to the weights. Precision levels can only increase during training and never decrease. Furthermore, MuPPET outputs a float32 network s.t. inference hast to be done in expensive full precision.

We advance the SOTA by providing an easy-to-use solution for quantized training of DNNs by using an information-theoretical intra-epoch precision switching mechanism capable of dynamically increasing and decreasing the precision of the network on a per-layer basis. Weights and activations are quantized to the lowest bit-width possible without information loss according to our heuristic and a certain degree of sparsity is induced while at the same time the bit-width is kept high enough for further learning steps to succeed. This results in a network which has advantages in terms of model size and time for training and inference. By training AlexNet and ResNet20 on the CIFAR10/100~datasets, we demonstrate on the basis of an analytical model for the computational cost that in comparison to a float32-baseline AdaPT is competitive in terms of accuracy and produces a non-trivial reduction of computational costs (speedup). Compared to MuPPET, AdaPT also has certain intrinsic methodological advantages. After AdaPT training, the model is fully quantized and sparsified to a certain degree s.t., unlike the case with MuPPET, which outputs a float32 model, AdaPT carries over it's advantages to the inference phase as well.

\section{Background}
\label{sec:bg}
\subsection{Quantization}
\label{ssec:bgquant}
\begin{figure}
\includegraphics[width=0.5\textwidth]{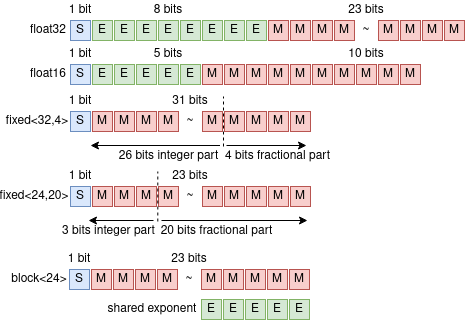}
\caption{\label{fig:numrep}\textit{Examples of floating-point, fixed-point and block-floating-point representations at different numerical precisions. Sign (S), Exponent (E), Mantissa (M) bits.}}
\end{figure}

Numerical representation describes how numbers are stored in memory (illustrated by fig. \ref{fig:numrep}) and how arithmetic operations on those numbers are conducted. Commonly available on consumer hardware are floating-point and integer representations while fixed-point or block-floating-point representations are used in high-performance ASICs or FPGAs. The numerical precision used by a given numerical representation refers to the amounts of bits allocated for the representation of a single number, e.g. a real number stored in float32 refers to floating-point representation in 32-bit precision. With these definitions of numerical representation and precision in mind, most generally speaking, quantization is the concept of running a computation or parts of a computation at reduced numerical precision or a different numerical representation with the intent of reducing computational costs and memory consumption. Quantized execution of a computation however can lead to the introduction of an error either through the quantized representations the machine epsilon $\epsilon_{mach}$ being too large (underflow) to accurately depict resulting real values or the representable range being too small to store the result (overflow).

\paragraph{Floating-Point Quantization}
\label{sssec:bgquantfloat}
The value $v$ of a floating point number is given by $v=\frac{s}{b^{p-1}} \times b^{e}$ where $s$ is the significand (mantissa), $p$ is the precision (number of digits in $s$), $b$ is the base and $e$ is the exponent~\cite{wilkinson1994roundingfl}. Hence quantization using floating-point representation can be achieved by reducing the number of bits available for mantissa and exponent, e.g. switching from a float32 to a float16 representation, and is offered out of the box by common machine learning frameworks for post-training quantization\cite{pytorch2019, tensorflow2015-whitepaper}.

\paragraph{Integer Quantization}
\label{sssec:bgquantint}
Integer representation is available for post-training quantization and QAT (int8, int16 due to availability on consumer hardware) in common machine learning frameworks~\cite{pytorch2019, tensorflow2015-whitepaper}. Quantized training however is not supported due to integer quantized activations being not meaningfully differentiable, making standard back-propagation inapplicable~\cite{yin2019understanding}. Special cases of integer quantization are 1-bit and 2-bit quantization, which are often referred to as binary and ternary quantization in literature.

\paragraph{Block-Floating-Point Quantization}
\label{sssec:bgquantbfp}
Block-floating-point represents each number as a pair of $WL$ (word length) bit signed integer $x$ and a scale factor $s$ s.t. the value $v$ is represented as $v = x \times b^{-s}$ with base $b=2$ or $b=10$. The scaling factor $s$ is shared across multiple variables (blocks), hence the name block-floating point, and is typically determined  s.t. the modulus of the larges element is $\in [\frac{1}{b}, 1]$~\cite{wilkinson1994roundingbfp}. Block-floating-point arithmetic is used in cases where variables cannot be expressed with sufficient accuracy on native fixed-point hardware.

\paragraph{Fixed-Point Quantization}
\label{sssec:bgquantfp}
Fixed-point numbers have a fixed number of decimal digits assigned and hence every computation must be framed s.t. the results lies withing the given boundaries of the representation~\cite{wilkinson1994roundingfp}. By definition of~\cite{hopkins2020stochastic}, a signed fixed-point numbers of world length $WL=i+s+1$ can be represented by a 3-tuple $\left \langle s,i,p \right \rangle$ where $s$ denotes whether the number is signed, $i$ denotes the number of integer bits and $p$ denotes the number of fractional bits.

\subsection{MuPPET}
\label{ssec:bgmuppet}
MuPPET is a mixed-precision DNN training algorithm that combines the use of block-floating and floating-point representations. The algorithm stores two copies of the networks weights: a float32 master copy of the weights that is updated during backwards passes and a quantized copy used for forward passes. 
MuPPETs uses block-floating point representation as outlined in sec. \ref{ssec:bgquant} with base $b=2$ for quantization. The precision level $i$ of a layer $l$ of all layers $\mathbb{L}$ is defined as 
\begin{equation}
     q^i_l = \left \langle WL^{net}, s_l^{weights}, s_l^{act} \right \rangle ^i
     \nonumber
\end{equation}
 with $WL^{net}$ being global across the network, and scaling factor $s$ (for weights and activation functions) varying per layer, determined each time precision switch is triggered. The scaling factor for a matrix $X$ is given by
\begin{dmath*} 
s^{weights,act} = 
\left \lfloor log_{2} 
min
\left ( 
\left ( 
\frac{UB+0.5}{\mathbf{X}_{max}^{\{weights,act\}}},
\frac{LB-0.5}{\mathbf{X}_{min}^{\{weights,act\}}}
\right )
\right ) 
\right \rfloor
\nonumber
\end{dmath*}
with $\mathbf{X}_{max,min}^{\{weights,act\}}$ describing the maximum or minimum value in the weights or feature maps matrix of the layer. $UB$ and $LB$ describe the upper bounds and lower bounds of the of the word length $WL^{net}$.
Its individual elements $x$ are quantized:
\begin{dmath*}
    x_{quant}^{\{weights,act\}} = \lfloor x^{\{weights,act\}} *  2^{s^{\{weights,act\}}} + \text{Unif}(-0.5,0.5) \rceil
    \nonumber
\end{dmath*}
$Unif(a,b)$ is the sampling from a uniform distribution in the interval $[a,b]$. The parameters $0.5$ and $-0.5$ to add to the bounds is chosen for maximum utilisation of $WL^{net}$. 
During training the precision of the quantized weights is increased using a precision-switching heuristic based on gradient diversity~\cite{gradientdiversity2018yin}, where the gradient of the last mini-batch $\nabla f_l^j(\boldsymbol{w}) $ of each layer $l \in \mathbb{L}$ at epoch $j$ is stored and after a certain number of epochs ($r$), the inter-epoch gradient diversity $\Delta s$ at epoch $j$ is computed by:
\begin{equation}
\Delta s(\boldsymbol{w})^j = \frac{\sum_{\forall l \in \mathbb{L}} \frac{\sum_{k = j-r}^j \| \nabla  f_l^k(\boldsymbol{w})\| ^2_2}{\|{\sum_{k = j-r}^j \nabla  f_l^k(\boldsymbol{w})\|^2_2}}}{\mathbb{L}} 
\nonumber
\end{equation}
At epoch $j$ there exists a set of gradient diversities $S(j) = \{ \Delta s(\boldsymbol{w})^i \forall e \leq i < j \}$ (e denotes the epoch in which it was switched into the quantization scheme) of which the ratio $p = \frac{\max S(j)}{\Delta s(\boldsymbol{w})^i}$ is calculated. If $p$ violates a threshold for a certain number of times, a precision switch is triggered. Speedups claimed by MuPPET are based on an estimated performance model simulating fixed-point arithmetic using NVIDIA CUTLASS~\cite{nvidia2020cutlass} for compatibility with GPUs, only supporting floating and integer arithmetic.

We chose MuPPET as baseline to compare AdaPT against because of the examined related work for the quantized training task, MuPPET comes closest to the goal of iso-accuracy.
\section{AdaPT}
\label{sec:marvin}
\subsection{Quantization Friendly Initialization}
\label{ssec:marvininit}
\begin{figure}[!h]
\includegraphics[width=0.5\textwidth]{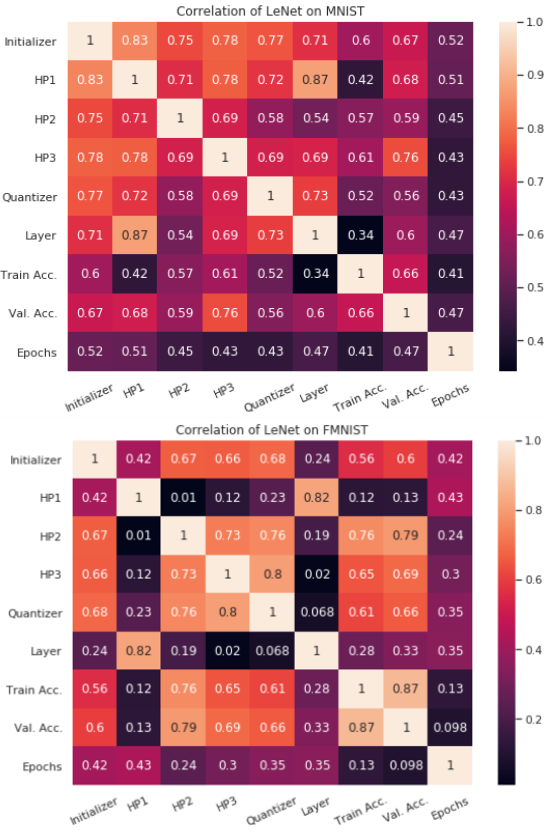}
\caption{\label{fig:preliminary}\textit{Correlation of initializer choice, hyperparameters (HP), quantizer, layer, training and validation accuracy, duration of training (epochs) for LeNet5 on MNIST and FMNIST datasets}}
\end{figure}

Despite exhaustive literature research and to our best knowledge, it has not yet been explored how weights initialization as counter-strategy for vanishing/exploding gradients impacts quantized training. However the preliminary experimental results of our first investigation of the impact of different weights initialization strategies showed the resilience of DNNs trained under a fixed forwards pass integer quantization scheme (int2, int4, int8, int16) with float32 master copies for gradient computations correlates strongly with initializer choice (fig. \ref{fig:preliminary}). Using Adam~\cite{kingma2014adam} as optimizer, we trained LeNet-5 on MNIST/FMNIST~\cite{mnist2012} and AlexNet on CIFAR10/100~\cite{cifar20XX} and examined the degree to which the quantized networks are inferior in accuracy compared to baseline networks trained in float32 dependent on initializer choice (Random Normal, Truncated Normal, Random Uniform, Glorot Normal/Uniform~\cite{glorot2010understanding}, He Normal/Uniform~\cite{he2015delving}, Variance Scaling~\cite{hanin2018start}, Lecun Normal/Uniform~\cite{klambauer2017self, lecun2012efficient}) and initializer parameters. We found that DNNs initialized by fan-in truncated normal variance scaling (TNVS) degrade least under quantized training with a fixed integer quantization scheme as described above. This correlates with results published by~\cite{bhalgat2020lsq} who introduced a learnable scale parameters in their asymmetric quantization scheme to achieve more stable training. We thus initialize networks with TNVS and an empirically chosen scaling factor $s$ where $n^l$ is the number of input units of a weights tensor $\boldsymbol{W}^l$ before quantized training with AdaPT.
\begin{equation}
    \boldsymbol{W}^l \sim  N \left ( \sigma=\sqrt{\frac{s}{n^l}},\mu=0,\alpha=\pm \sqrt{\frac{3\cdot s}{n^l}} \right )
    \nonumber
\end{equation} 
The examination of how other counter strategies (eg. gradient amplification,~\cite{basodi2020gradient}, gradient normalization~\cite{chen2018gradnorm}, weights normalization~\cite{salimans2016weight}) affect quantized training remains an open question.
\subsection{Precision Levels}
AdaPT uses fixed-point quantization as defined in sec. \ref{sssec:bgquantfp} because unlike block floating point as used by MuPPET with a global word length across the whole network and per-layer shared exponent, we conjecture that optimal word and fractional lengths are local properties of each layer under the hypothesis that different layers contain different amounts of information during different points of time in training. We decided against using floating-point quantization because fixed-point gives us better control over numerical precision and is available in high performance ASICs which are our target platform and against integer quantization, particularly its extreme forms binarization and ternarization, because our goal is iso-accuracy which literature shows is difficult to reach with such coarse quantization. In principle however, the AdaPT concept could be extended to representations other than fixed-point. Furthermore, AdaPT employs stochastic rounding, which in combination with a fixed-point representation has been shown to consistently outperform nearest-rounding by~\cite{hopkins2020stochastic}, for quantizing float32 numbers. Given that AdaPT is agnostic towards whether a number is signed or not, we represent the precision level of each $l \in \mathbb{L}$ simply as $\left \langle WL^l, FL^l \right \rangle$ whereby fractional length $FL^l$ denotes the number of fractional bits. For a random number $P \in [0,1]$, $x$ is stochastically rounded by
\begin{equation}SR(x)=\left\{
\begin{array}{ll}
\left \lfloor x \right \rfloor, \text{if } P \geq \frac{x-\left \lfloor x \right \rfloor}{\epsilon_{mach}}\\
\left \lfloor x \right \rfloor+1, \text{if } P < \frac{x-\left \lfloor x \right \rfloor}{\epsilon_{mach}}\\
\end{array}
\right.
\nonumber
\end{equation}

\subsection{Precision Switching Mechanism}
\label{ssec:marvinswitch}
Precision switching in quantized DNN training is the task of carefully balancing the need to keep precision as low as possible in order to improve runtime and model size, yet still keep enough precision for the network to keep learning. In AdaPT, we have encoded these opposing interests in two operations, the PushDown Operation and the PushUp Operation. 
\paragraph{The PushDown Operation}
Determining the amount of information lost if the precision of the fixed-point representation of a layer's weight tensor is lowered, can be heuristically accomplished by interpreting the precision switch as a change of encoding. Assume a weights tensor $\boldsymbol{W}^l \sim Q^l$ of layer $l \in \mathbb{L}$ with its quantized counterpart $\widehat{\boldsymbol{W}}^l \sim P^l$, where $P^l, Q^l$ are the respective distributions. Then the continuous Kullback-Leibler-Divergence~\cite{kullback1951information} \eqref{eq:ckld} represents the average number of bits lost through changing the encoding of $l$ from $Q^l$ to $P^l$, with $p$ and $q$ denoting probabilities, and $w$ the elements of the weights tensor:
\begin{equation}
   D(P^l\|Q^l)=\int_{ -\infty}^{ \infty}p^l(w) \cdot log \frac{p^l(w)}{q^l(w)}dw 
   \label{eq:ckld}
   \nonumber
\end{equation}
Using discretization via binning, we obtain $P^l$ and $Q^l$ at resolution $r^l$ through the empirical distribution function: 
\begin{equation}
\widehat{F_{r^l}}(w)=\frac{1}{r^l+1}\sum_{i=1}^{r^l}1_{W_i^l\leq w}
\label{eq:edf}
\end{equation}
that can then be used in the discrete Kullback-Leibler-Divergence \eqref{eq:dkld}.
\begin{equation}
   KL(P^l\|Q^l)=\sum_{w \in \boldsymbol{W}^l}P^l(w)\cdot log \frac{P^l(w)}{Q^l(w)}
   \label{eq:dkld}
\end{equation}
Using a bisection approach, AdaPT efficiently finds the smallest quantization $\left \langle WL_{min}^l, FL_{min}^l \right \rangle$ of $\boldsymbol{W}^l$ s.t. $KL(P^l\|Q^l) = 0$ $\forall l \in \mathbb{L}$
\paragraph{The PushUp Operation}
However, determining the precision of the fixed-point representation of $\boldsymbol{W}^l$ at batch $j$, s.t. the information lost through quantization is minimal but there is still sufficient precision for subsequent batches $j+1$ to keep learning, is a non-trivial task. Solely quantizing $\boldsymbol{W}^l$ at the beginning of the training to a low precision fixed-point representation (e.g. $\left \langle WL^l,FL^l \right \rangle = \left \langle 8,4 \right \rangle$) would result in the network failing to learn, because at such low precision levels, gradients would vanish very early on in the backwards pass. Hence, AdaPT tracks for each layer a second heuristic over the last $j$ batches to determine how much precision is required for the network to keep learning. If gradients are quantized, a gradient diversity based heuristic is employed \eqref{eq:ourgd}, \eqref{eq:loggd}.
\begin{equation}
\Delta s(\boldsymbol{w})^l_j = \frac{\sum_{k = j-r}^j \| \nabla  f_k^l(\boldsymbol{w})\| _2}{\|{\sum_{k = j-r}^j \nabla  f_k^l(\boldsymbol{w})\|_2}}
\label{eq:ourgd}
\end{equation}
\begin{equation}
    \Delta \tilde{s}(\boldsymbol{w})^l_j =
    \begin{cases}
      log \Delta s(\boldsymbol{w})^l_j & \text{if $0 < \Delta s(\boldsymbol{w})^l_j < \infty$}\\
      1 & \text{otherwise}\\
    \end{cases} 
    \label{eq:loggd}
    \nonumber
\end{equation}
If $\Delta \tilde{s}(\boldsymbol{w})^l_j > 0$, two suggestions for an increase in precision are computed, $s_1^l = max(\lceil\frac{1}{log\Delta s(\boldsymbol{w})^l_j-1}\rceil, 1)$ and $s_2^l = max(min(32 \cdot log^2\Delta s(\boldsymbol{w})^l_j-1, 32) - FL_{min}^l,1)$ and the final suggestion is computed dependent on a global strategy $st$ via \begin{equation}
    s^l =
    \begin{cases}
     min(s_1^l,s_2^l) & \text{if st = min}\\
     \lceil 0.5 \cdot (s_1^l+s_2^l) \rceil & \text{if st = mean}\\
     max(s_1^l,s_2^l) & \text{if st = max}\\
    \end{cases}
    \label{eq:pusuggestionfinal}
\end{equation}
Otherwise, i.e. $\Delta \tilde{s}(\boldsymbol{w})^l_j > 0$, $s^l=1$. The new fixed-point quantization of layer $l$ is then obtained by $FL^l = (min(FL_{min}^l + s^l, 32)$, $WL^l = min(max(WL_{min}^l, FL_{min}^l)+1, 32))$.
\paragraph{Dealing with Fixed-Points Limited Range}
As outlined in sec. \ref{sssec:bgquantfp}, fixed-point computations must be framed s.t. results fit within the given boundaries. We approach this by adding a number of buffer bits $buff$ to every layers word-length, i.e. at the end of PushUp, $FL^l = (min(FL_{min}^l, 32-buff)$, $WL^l = max(min( FL_{min}^l+buff, 32),WL_{min}^l)$. Additionally, we normalize gradients to limit weight growth and reduce chances of weights weights becoming unrepresentable in the given precision after an update step.
\begin{equation}
     \nabla  f^l(\boldsymbol{w}) = \frac{\nabla  f^l(\boldsymbol{w})}{\| \nabla  f^l(\boldsymbol{w})\| _2}
     \nonumber
\end{equation}

\paragraph{Strategy, Resolution and Lookback}
For adapting the strategy $st$ mentioned in \eqref{eq:pusuggestionfinal}, we employ a simple loss-based heuristic. First we compute the average lookback over all layers $lb_{avg} = |\mathbb{L}|^{-1} \sum_{i=0}^{|\mathbb{L}|} lb^l$ and average loss $\mathcal{L}_{avg} = |\mathcal{L}|^{-1} \sum_{i=0}^{lb_{avg}} \mathcal{L}_{i}$ over the last $lb_{avg} $ batches. Then via \eqref{eq:stratadapt}, the strategy is adapted.
\begin{equation}
    st =
\begin{cases}
 \text{max} & \text{if $|\mathcal{L}_{avg}| \leq |\mathcal{L}_{i}|$ and st = 'mean'}\\
 \text{mean} & \text{if $|\mathcal{L}_{avg}| \leq |\mathcal{L}_{i}|$ and st = 'min'}\\
 \text{min} & \text{if $|\mathcal{L}_{avg}| > |\mathcal{L}_{j}|$}\\
\end{cases}
\label{eq:stratadapt}
\nonumber
\end{equation}
Because the number of gradients collected for each layer affects the result of the gradient diversity based heuristic \eqref{eq:ourgd}, \eqref{eq:loggd}, we introduce a parameter lookback $lb^l$ bounded by hyperparameters $lb_{lwr}  \leq lb^l \leq lb_{upr}$ which is estimated at runtime. First, $lb_{new}$ is computed: 
\begin{dmath*}
    lb_{new}^l =
    \begin{cases}
      min(max(\lceil \frac{ lb_{upr}}{\Delta s(\boldsymbol{w})^l_j} \rceil, lb_{lwr}), lb_{upr})\ & \text{if $0<\Delta s(\boldsymbol{w})^l_j$}\\
      lb_{upr} & \text{otherwise}
    \end{cases} 
    \label{eq:llbnew}
    \nonumber
\end{dmath*}
Then, to prevent jitter, a simple momentum is applied to obtain the updated $lb^l = \lceil lb_{new}^l \cdot \gamma + (1-\gamma) \cdot lb^l \rceil$ with $\gamma \in [0,1]$. \\
Similarly, the number of bins used in the discretization step \eqref{eq:edf} affects the result of the discrete Kullback-Leibler-Divergence \eqref{eq:dkld}. We control the number of bins via a parameter referred to as resolution $r^l$, which is derived at runtime and bounded by hyperparameters $r_{lwr}  \leq r^l \leq r_{upr}$. 
\begin{equation}
    r^l =
    \begin{cases}
      min(max(r^l+1, r_{lwr}), r_{upr}) & \text{if $lb^l = lb_{upr}$}\\
      min(max(r^l-1, r_{lwr}), r_{upr}) & \text{if $lb^l = lb_{lwr}$}
    \end{cases} 
    \label{eq:rlcontrol}
\end{equation}






\subsection{AdaPT-SGD (ASGD)}
\label{ssec:marvinsgd}
Although AdaPT can in principle be combined with any iterative gradient based optimizer (e.g. Adam), we chose to implement AdaPT with Stochastic Gradient Descent (SGD), because it generalizes better than adaptive gradient algorithms~\cite{NEURIPS2020_f3f27a32}. The implementation of AdaPT employs the precision switching mechanism and numeric representation described in sec. \ref{sec:marvin}, by splitting it in two operations: the PushDown Operation (alg. \ref{alg:pdop}), which, for a given layer, finds the smallest fixed-point representation that causes no information loss, and the PushUp Operation (alg. \ref{alg:puop}), which, for a given layer, seeks the precision that is required for the network to keep learning. The integration of these two operations into the precision switching mechanism is depicted in alg. \ref{alg:precswitch}, and the integration into SGD training process is depicted in alg. \ref{alg:marvinsgd}. 
\paragraph{Inducing Sparsity}
In addition to the AdaPT precision switching mechanism, we used an L1 sparsifying regularizer~\cite{williams1995bayesian, ng2004feature, tibshirani1996regression} to obtain sparse and particularly quantization-friendly weight tensors, combined linearly with L2 regularization for better accuracy in a similar way as proposed by~\cite{zou2005regularization}. Additionally using a technique similar to~\cite{morphnet2018}, we penalize learning steps that lead to increased word-length or decreased sparsity by:
\begin{equation}
    \mathcal{P} = \frac{WL^l}{32}\cdot sp^l
    \label{eq:optim1}
    \nonumber
\end{equation}
where $sp^l$ is the percentage of non-zero elements of layer $l$. Thus loss in ASGD is computed by:
\begin{equation}
    \widehat{\mathcal{L}}(\boldsymbol{W}^l)= \mathcal{L} + \alpha\left \| \boldsymbol{W}^l \right \|_1 + \frac{\beta}{2}\left \| \boldsymbol{W}^l \right \|^2_2+\mathcal{P}
    \label{eq:optim2}
    \nonumber
\end{equation}
\paragraph{Algorithm}
\begin{algorithm}
\LinesNumbered
\DontPrintSemicolon
 \KwData{Untrained Float32 DNN}
 \KwResult{Trained and Quantized DNN}
 $\widehat{\mathbb{L}}$ = InitFixedPointTNVS()\;
 $\mathbb{Q}$ = InitQuantizationMapping($\widehat{\mathbb{L}}$)\;
 $\mathbb{L}$ = Float32Copy($\widehat{\mathbb{L}}$)\;
 \For{Epoch in Epochs}{
     \For{Batch in Batches}{
        $\widehat{\mathbb{G}}$, $\mathcal{L}$ = ForwardPass($\widehat{\mathbb{L}}$, $Batch$)\;
        $\mathbb{Q}$ = PrecisionSwitch($\widehat{\mathbb{G}}$, $\mathcal{L}$, $\mathbb{Q}$, $\mathbb{L}$)\;
        SGDBackwardsPass($\mathbb{L}$, $\widehat{\mathbb{G}}$)\;
        \For{$l \in \mathbb{L}$} {
        $\widehat{\mathbb{L}}[l]$ = Quantize($\mathbb{L}[l]$, $\mathbb{Q}[l][quant]$)\;
        }
     }
 }
 \caption{AdaPT-SGD}
 \label{alg:marvinsgd}
\end{algorithm}
When AdaPT-SGD (alg. \ref{alg:marvinsgd}) is started on an untrained float32 DNN, it first initializes the DNNs layers (denoted as $\widehat{\mathbb{L}}$) weights with TNVS (alg. \ref{alg:marvinsgd}, ln. 1). Next the quantization mapping $\mathbb{Q}$ is initialized, which assigns each $l \in \mathbb{L}$ a tuple $\left \langle WL^l,FL^l \right \rangle$, a lookback $lb^l$ and a resolution $r^l$ (alg. \ref{alg:marvinsgd}, ln. 2). Then a float32 master copy $\mathbb{L}$ of $\widehat{\mathbb{L}}$ is created (alg. \ref{alg:marvinsgd}, ln. 3). During training for each forward pass on a batch, quantized gradients $\widehat{\mathbb{G}}$ and loss $\mathcal{L}$ are computed with a forward pass using quantized layers $\widehat{\mathbb{L}}$ (alg. \ref{alg:marvinsgd}, ln. 4-6).
\begin{algorithm}
\LinesNumbered
\DontPrintSemicolon
 \KwData{$\widehat{\mathbb{G}}$, $\mathcal{L}, \mathbb{Q}, \mathbb{L}$}
 \KwResult{$\mathbb{Q}$}
    AdaptStrategy($\mathcal{L}$)\;
    \For{$l \in {\mathbb{L}}$} {
        $\mathbb{Q}[l][grads].append(\widehat{\mathbb{G}}[l])$\;
        AdaptLookback($\mathbb{Q}[l][grads], \mathbb{Q}[l][lb]$)\;
        AdaptResolution($\mathbb{Q}[l][lb], \mathbb{Q}[l][res]$)\;
        \If{$|\mathbb{Q}[l][grads]| \geq \mathbb{Q}[l][lb]$}{$WL^l$, $FL^l$ = $\mathbb{Q}[l][quant]$\;
        $WL^l_{min}$, $FL^l_{min}$ = PushDown($\mathbb{L}[l]$, $WL^l$, $FL^l$)\;
        $WL^l_{new}$, $FL^l_{new}$ = PushUp($l$, $WL^l$, $FL^l$, $WL^l_{min}$, $FL^l_{min}$)\;
        $\mathbb{Q}[l][quant]$ = $WL^l_{new}$, $FL^l_{new}$\;}
 }
 \caption{PrecisionSwitch}
 \label{alg:precswitch}
\end{algorithm}
The precision switching mechanism described in sec. \ref{sec:marvin} is then called (alg. \ref{alg:marvinsgd}, ln. 7) and after adapting the push up strategy as described in sec. \ref{sec:marvin}, it iterates over $l \in \mathbb{L}$ (alg. \ref{alg:precswitch} ln. 2), first adapting resolution $r^l$ and lookback $lb^l$ (alg. \ref{alg:precswitch} ln. 4-5) and then executing PushDown and PushUp on layer $l$  (alg. \ref{alg:precswitch} ln. 6 -10) to update $\left \langle WL^l,FL^l \right \rangle \in \mathbb{Q}$.
\begin{algorithm}
\LinesNumbered
\DontPrintSemicolon
 \KwData{$L$, $WL^l$, $FL^l$}
 \KwResult{$WL^l_{min}$, $FL^l_{min}$}
    $WL^l_{min}$, $FL^l_{min}$ = $L$, $WL^l$, $FL^l$\;
    \Repeat{$KL(EDF(L_i), EDF(\widehat{L})) < \epsilon$}{
        $WL^l_{min}$, $FL^l_{min}$ = Decrease($WL^l_{min}$, $FL^l_{min}$)\;
        $\widehat{L}$ = Quantize($L$, $WL^l_{min}$, $FL^l_{min}$)\;
 }
 \caption{PushDown Operation}
  \label{alg:pdop}
\end{algorithm}
When PushDown is called on $l$, it decreases the quantization mapping in a bisectional fashion until $KL$ indicates amore coarse quantization would cause information loss at resolution $r^l$ (\ref{alg:pdop}, ln. 1-5).
\begin{algorithm}
\LinesNumbered
\DontPrintSemicolon
 \KwData{$L$, $WL^l$, $FL^l$, $WL^l_{min}$, $FL^l_{min}$}
 \KwResult{$WL^l_{new}$, $FL^l_{new}$}
    Compute $\widehat{\Delta s(L)}$ using $WL^l_{min}$, $WL^l_{min}$\;
    Compute $\Delta s(L)$ using $WL^l$, $FL^l$\;
    \While{$\Delta s(L)$ $\geq$ $\widehat{\Delta s(L)}$}{
        $WL^l_{min}$, $FL^l_{min}$ = Increase($WL^l_{min}$, $FL^l_{min}$)\;
        Compute $\widehat{\Delta s(L)}$ using $WL^l_{min}$, $FL^l_{min}$\;
 }
 \caption{PushUp Operation}
 \label{alg:puop}
\end{algorithm}
After computing the most coarse quantization $\left \langle WL^l_min,FL^l_min \right \rangle \in \mathbb{Q}$ not causing information loss in $l$, PushUp is called on $l$ to increase the quantization to the point where the network is expected to keep learning, based on gradient diversity of the last $lb^l$ batches (alg. \ref{alg:puop}, ln. 1-6). After PushUp the PrecisionSwitch returns an updated $\mathbb{Q}$ to the training loop, and a regular SGD backward pass updates the float32 master copy $\mathbb{L}$, using quantized gradients $\widehat{\mathbb{G}}$ (alg. \ref{alg:marvinsgd}, ln. 7,8). Finally, the now updated weights $\mathbb{L}$ are quantized using the updated $\mathbb{Q}$ and written back to $\widehat{\mathbb{L}}$ to be using in the next forward pass (alg. \ref{alg:marvinsgd}, ln. 9-11).
\section{Experimental Evaluation}
\label{sec:eval}
\subsection{Setup}
\label{ssec:setup}
For experimental evaluation of AdaPT, we trained AlexNet and ResNet20 on the CIFAR-10/100 datasets with reduce on plateau learning rate (ROP) scheduling which will reduce learning rate by a given factor if loss has not decreased for a given number of epochs~\cite{pytorch2019}. Due to unavailability of fixed-point hardware, we used QPyTorch to simulate fixed point arithmetic and our own performance model (sec. \ref{sssec:permo}) to simulate speedups, model size reductions and memory consumption. Experiments were conducted on an Nvidia DGX-1. It has 8 x Tesla V100 GPUs, which is capable of integer and floating-point arithmetic~\cite{nvidiaDGX1datasheet}.
\subsubsection{Hyper Parameters}
All layers $l \in \mathbb{L}$ were quantized with $\left \langle WL^l, FL^l \right \rangle = \left \langle 8, 4 \right \rangle$ at the beginning of AdaPT training. Other hyperparameters specific to AdaPT were set to $r_{lwr}=50$, $r_{upr}=150$, $lb_{lwr}=25$, $lb_{upr}=100$, lookback momentum $\gamma = 0.33$ for all experiments, buffer bits was set to $buff=4$ for training AlexNet on CIFAR10 and $buff=8$ for training ResNet and AlexNet on CIFAR100.  Hyperparameters unspecific to AdaPT ($lr$, $L1_{decay}$, $L2_{decay}$, $ROP_{patience}$, $ROP_{threshold}$ \textit{batch size}, \textit{accumulation steps}) were selected using grid search and 10-fold cross-validation and we refer to our code repository \footnote{\url{https://gitlab.cs.univie.ac.at/sidakk95cs/marvin2}} for the exact configuration files of each experiment.
\subsubsection{Performance Model}
\label{sssec:permo}
Speedups and model size reductions were computed using a performance model taking forward and backward passes, as well as batch sizes, gradient accumulations numerical precision and sparsity, into account. Our performance model computes per layer operations (MAdds, subsequently referred to as ops), weights them with the layer's world length and a tensors percentage of non-zero elements at a specific stage in training to simulate quantization and a sparse tensor format, and aggregates them to obtain the overall incurred computational costs of all forwards and backwards passes. Additionally, the performance model estimates AdaPTs overhead for each layers $l$ push up operation $pu^l$ and push down operation $pd^l$ by 
\begin{equation}
    ops^l_{pd, i} \leq 2 \cdot log_{2} (32-8) \cdot r^l_{i} \cdot 3 \cdot \prod_{dim \in l}dim
    \label{eq:pdops}
\end{equation}
\begin{equation}
    ops^l_{pu, i} \leq (lb_{i}+1) \cdot \prod_{dim \in l}dim+1
    \label{eq:puops}
\end{equation}
Using \eqref{eq:pdops}, \eqref{eq:puops} and the simplifying assumption that a backwards pass incurs as many operations as a quantized forwards pass but is conducted in full precision i.e. 32 bits word length with non-sparse gradients, AdaPTs training costs are then bounded by
\begin{equation}
    costs_{train} \leq \sum_{i=1}^n \sum_{l=1}^{|\mathbb{L}|} ops^l \cdot \left ( sp^l_i \cdot  WL_i^l + \frac{32}{accs} \right)
    \label{eq:costs}
\end{equation}
and AdaPTs overhead is bounded by
\begin{equation}
    costs_{AdaPT} \leq \sum_{i=1}^n \sum_{l=1}^{|\mathbb{L}|} 32 \cdot \frac{sp^l_i \cdot ops^l_{pd,i} + ops^l_{pu,i}}{ accs \cdot lb^l_i} 
    \label{eq:costs_oh}
\end{equation}
where $n$ is the number of training steps, $\mathbb{L}$ are the networks layers and $WL_i^l$ is the $l-ths$ layers word length at training step $i$ and $sp^l_i$ is the percentage of non-zero elements of layer $l$ at step $i$. Using \eqref{eq:costs}, \eqref{eq:costs_oh}, we obtain total costs via $costs = cost_{train} + costs_{AdaPT}$ and further the speedup of our training approach compared to MuPPET or a float32 baseline incorporating batch size $bs$, gradient accumulation steps $accs$ and computational costs $costs$ is thus obtained via
\begin{equation}
    SU = \frac{bs_{other} \cdot costs_{other}} {bs_{ours} \cdot costs_{ours}}
    \nonumber
\end{equation}
whereby naturally we exclude AdaPTs overhead when computing $costs_{other}$.
Models size reductions $SZ$ were calculated by first computing individual model sizes $sz$
\begin{equation}
    sz = \sum_{l=1}^{|\mathbb{L}|} sp^l_i \cdot WL_i^l
    \nonumber
\end{equation} with $i = n$ and then forming the quotient $SZ = sz_{other}/sz_{ours}$. Similarly, average memory consumption during training $MEM=mem_{other}/mem_{ours}$ is formed by the quotient of each models memory consumption $mem$
\begin{equation}
    mem = \left ( \sum_{i=1}^n\sum_{l=1}^{|\mathbb{L}|} \left (sp^l_i \cdot WL_i^l+32 \right ) \right ) \cdot \frac{1}{n}
    \nonumber
\end{equation}
Because $sz$ and $mem$ ignore tensor dimensions, they can not be interpreted as absolute values. However given that in the quotients $MEM$ and $SZ$ the effects of tensor dimensions would cancel out when comparing identical architectures (i.e. float32 AlexNet and quantized AlexNet), these values provide valid relative measures for memory consumption and model size under this constraint.
\subsection{Results}
\label{ssec:results}
\subsubsection{Training}
\begin{figure}[!h]
\includegraphics[width=0.5\textwidth]{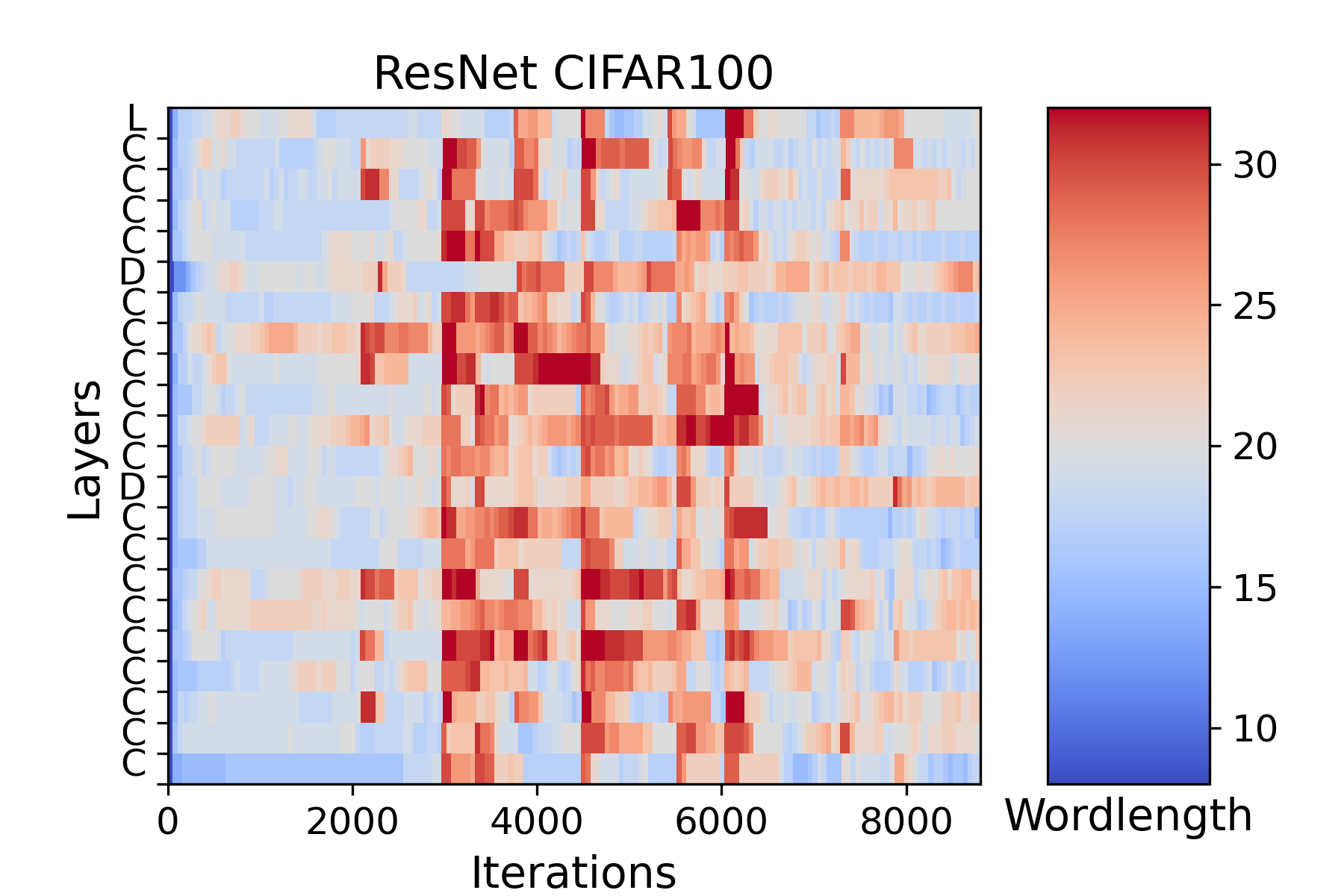}
\caption{\label{fig:eval_resnet20_c100_wl}\textit{Wordlengths (bit) of ASGD optimized ResNet20 on CIFAR100, 100 epochs, Linear (L), Convolutional (C), Downsampling (D) layers}}
\end{figure}

\begin{figure}[!h]
\includegraphics[width=0.5\textwidth]{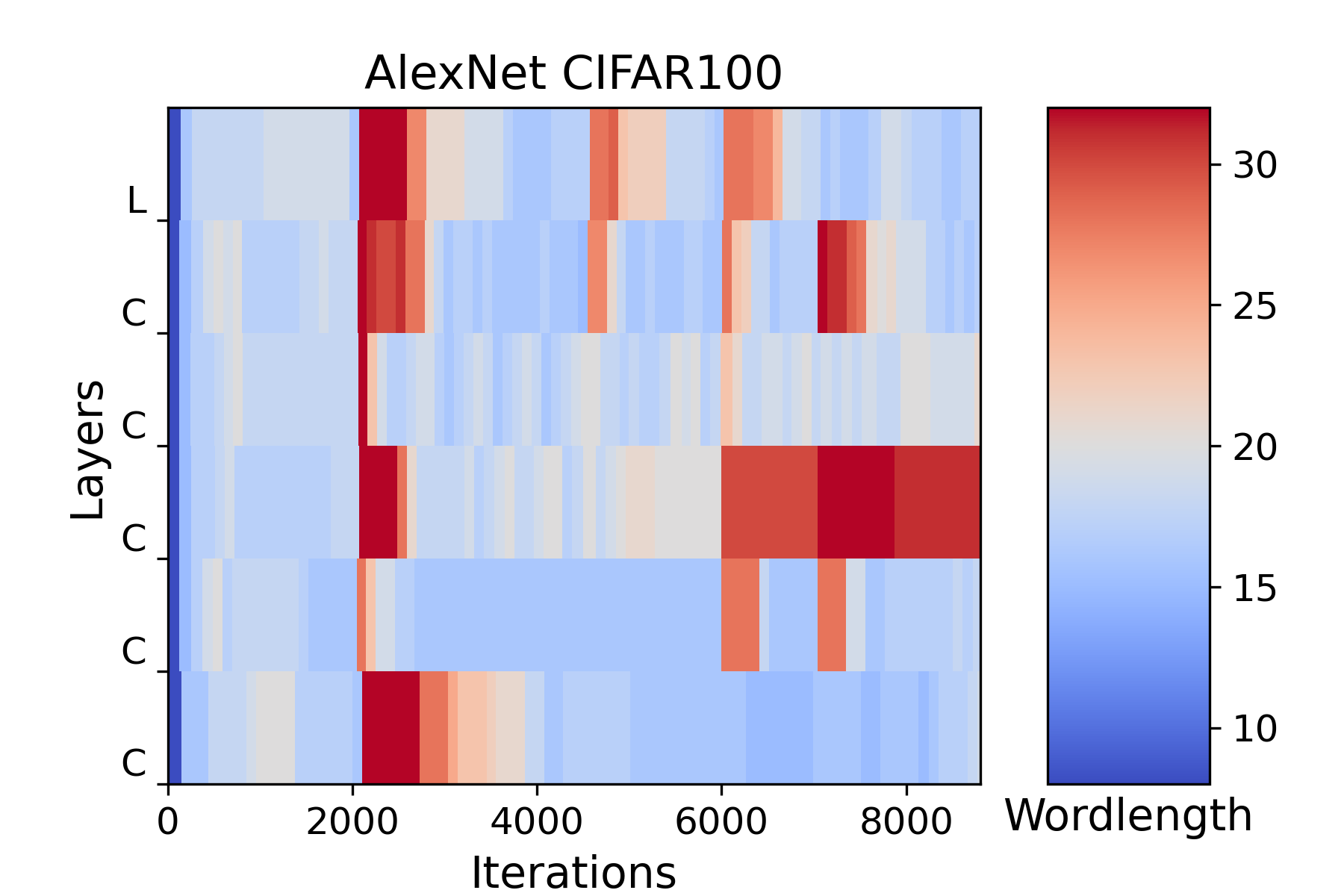}
\caption{\label{fig:eval_alexnet_c100_wl}\textit{Wordlengths (bit) of ASGD optimized AlexNet on CIFAR100, 100 epochs}}
\end{figure}

\begin{figure}[!h]
\includegraphics[width=0.5\textwidth]{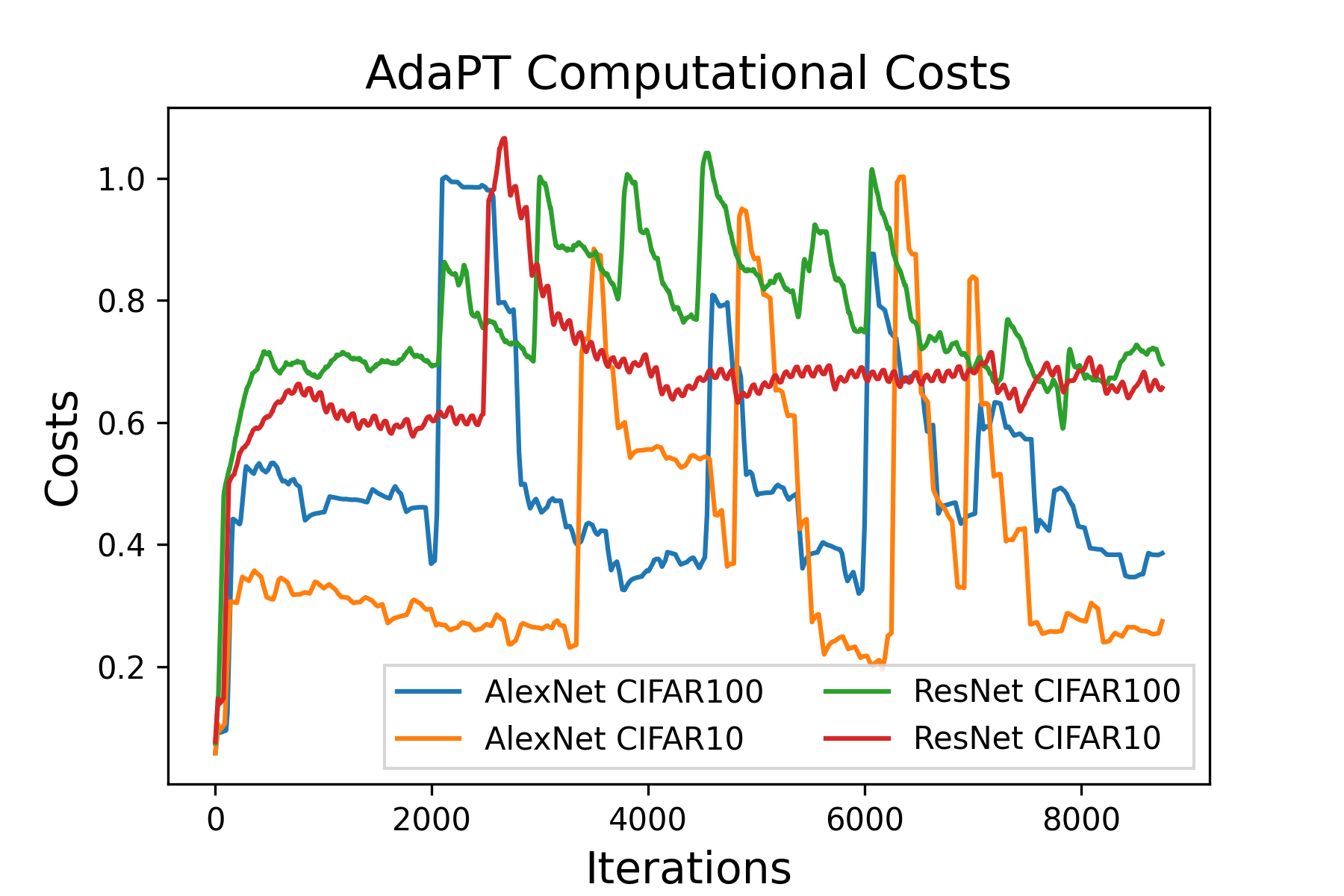}
\caption{\label{fig:asgd_costs}\textit{ASGD computational cost relative to float32 SGD}}
\end{figure}

Tab. \ref{tab:overview1} shows the top-1 validation accuracies and accuracy differences achieved by AdaPT and MuPPET for both quantized training on CIFAR100, and a float32 basis. Tab. \ref{tab:overview2} shows results for training on CIFAR10. As can be seen in the tables, AdaPT is capable of quantized training not only to an accuracy comparable to float32 training but even reaches or surpasses iso-accuracy in all examined cases. The chosen experiments furthermore illustrate that AdaPT delivers this performance independent of the underlying DNN architecture or the dataset used. 
\begin{table}[!h]
\centering
\begin{tabular}{|l|lll|lll|}
\hline
\textbf{CIFAR100} & \textbf{} & \textbf{} & \textbf{}\\
\textbf{} & \textbf{Float32} & \textbf{Quantized} & \textbf{$\Delta$} \\ \hline
\textbf{AlexNet}\textsubscript{AdaPT} & 41.3 $_{512}^{100}$ & 42.4 $_{512}^{100}$ & 1.1 \\ \hline
\textbf{AlexNet}\textsubscript{MuPPET} & 39.2 $_{128}^{150}$ & 38.2 $_{128}^{?}$& -1.0 \\ \hline
\textbf{ResNet}\textsubscript{AdaPT} & 64.3 $_{512}^{100}$ & 65.2 $_{512}^{100}$ & 0.9 \\ \hline
\textbf{ResNet}\textsubscript{MuPPET} & 64.6 $_{128}^{150}$ & 65.8 $_{128}^{?}$ & 1.2 \\ \hline
\end{tabular}
\caption{\label{tab:overview1} Top-1 accuracies, AdaPT (ours) vs MuPPET, CIFAR10, 100 to 150 epochs. Float 32 indicates 32-bit floating-point training (baseline), Quantized indicates variable bit fix-point (AdaPT) or block-floating-point (MuPPET) quantized training, subscript indicates batch size used, superscript indicates epochs used}
\end{table}
\begin{table}[!h]
\centering
\begin{tabular}{|l|lll|lll|}
\hline
\textbf{CIFAR10} & \textbf{} & \textbf{} & \textbf{}\\
\textbf{} & \textbf{Float32} & \textbf{Quantized} & \textbf{$\Delta$} \\ \hline
\textbf{AlexNet}\textsubscript{AdaPT} & 73.1 $_{512}^{100}$ & 74.5 $_{512}^{100}$ & 1.4 \\ \hline
\textbf{AlexNet}\textsubscript{MuPPET} & 75.5 $_{128}^{150}$ & 74.5 $_{128}^{99}$ & -1.0 \\ \hline
\textbf{ResNet}\textsubscript{AdaPT} & 89.5 $_{512}^{100}$ & 90.0 $_{512}^{100}$ & 0.5 \\ \hline
\textbf{ResNet}\textsubscript{MuPPET} & 90.1 $_{128}^{150}$ & 90.9 $_{128}^{114}$ & 0.8 \\ \hline
\end{tabular}
\caption{\label{tab:overview2} Top-1 accuracies, AdaPT (ours) vs MuPPET, CIFAR100}
\end{table}
Fig. \ref{fig:eval_resnet20_c100_wl} and \ref{fig:eval_alexnet_c100_wl} display word length usage over time for individual layers for ResNet20 and AlexNet on CIFAR100. In both cases, we observe that individual layers have different precision preferences dependent on progression of the training, an effect that is particularly pronounced in AlexNet. For ResNet20, the word length interestingly decreased notably in the middle of training and later decreased, which we conjecture is attributable to sparsifying L1 regularization as well as the wordlength/sparsity penalty introduced in sec. \ref{sec:marvin} leading to a reduction in irrelevant weights, that would otherwise offset the KL heuristic used in the PushDown Operation.
\begin{table}[!h]
\centering
\begin{tabular}{|l|llll|}
\hline
\textbf{CIFAR10} & \textbf{} & \textbf{} & \textbf{} & \\
& \textbf{MEM} & \textbf{SU}\textsuperscript{1} & \textbf{SU}\textsuperscript{2} & \textbf{SU}\textsuperscript{3} \\ \hline
\textbf{AlexNet}\textsubscript{AdaPT} & 1.47 & 1.42 & 2.37 & 6.42\\ \hline
\textbf{AlexNet}\textsubscript{MuPPET} & 1.66 & ? & ? & 1.2\\ \hline
\textbf{ResNet}\textsubscript{AdaPT} & 1.63 & 1.20 & 1.49 & 4.01\\ \hline
\textbf{ResNet}\textsubscript{MuPPET} & 1.61 & ? & ? & 1.25 \\ \hline
\end{tabular}
\caption{\label{tab:overview4} Memory Footprint, Final Model Size, Memory Footprint, Speedup, AdaPT (ours) vs MuPPET on respective baseline float32 training on CIFAR10. SU\textsuperscript{1}: our baseline, our performance model, SU\textsuperscript{2}: our baseline, our performance model adjusted for iso-accuracy, SU\textsuperscript{3}: MuPPET baseline, our performance model}
\end{table}
\begin{table}[!h]
\centering
\begin{tabular}{|l|llll|}
\hline
\textbf{CIFAR100} & \textbf{} & \textbf{} & \textbf{} & \\
& \textbf{MEM} & \textbf{SU}\textsuperscript{1} & \textbf{SU}\textsuperscript{2} & \textbf{SU}\textsuperscript{3} \\ \hline
\textbf{AlexNet}\textsubscript{AdaPT} & 1.41 & 1.32 & 1.57 & 5.23\\ 
\textbf{ResNet}\textsubscript{AdaPT} & 1.66 & 1.13 & 1.62 & 3.36 \\ \hline
\end{tabular}
\caption{\label{tab:overview3} Memory Footprint, Final Model Size, Memory Footprint, Speedup, AdaPT (ours) vs MuPPET, CIFAR100, training}
\end{table}
\begin{figure}[!h]
\includegraphics[width=0.5\textwidth]{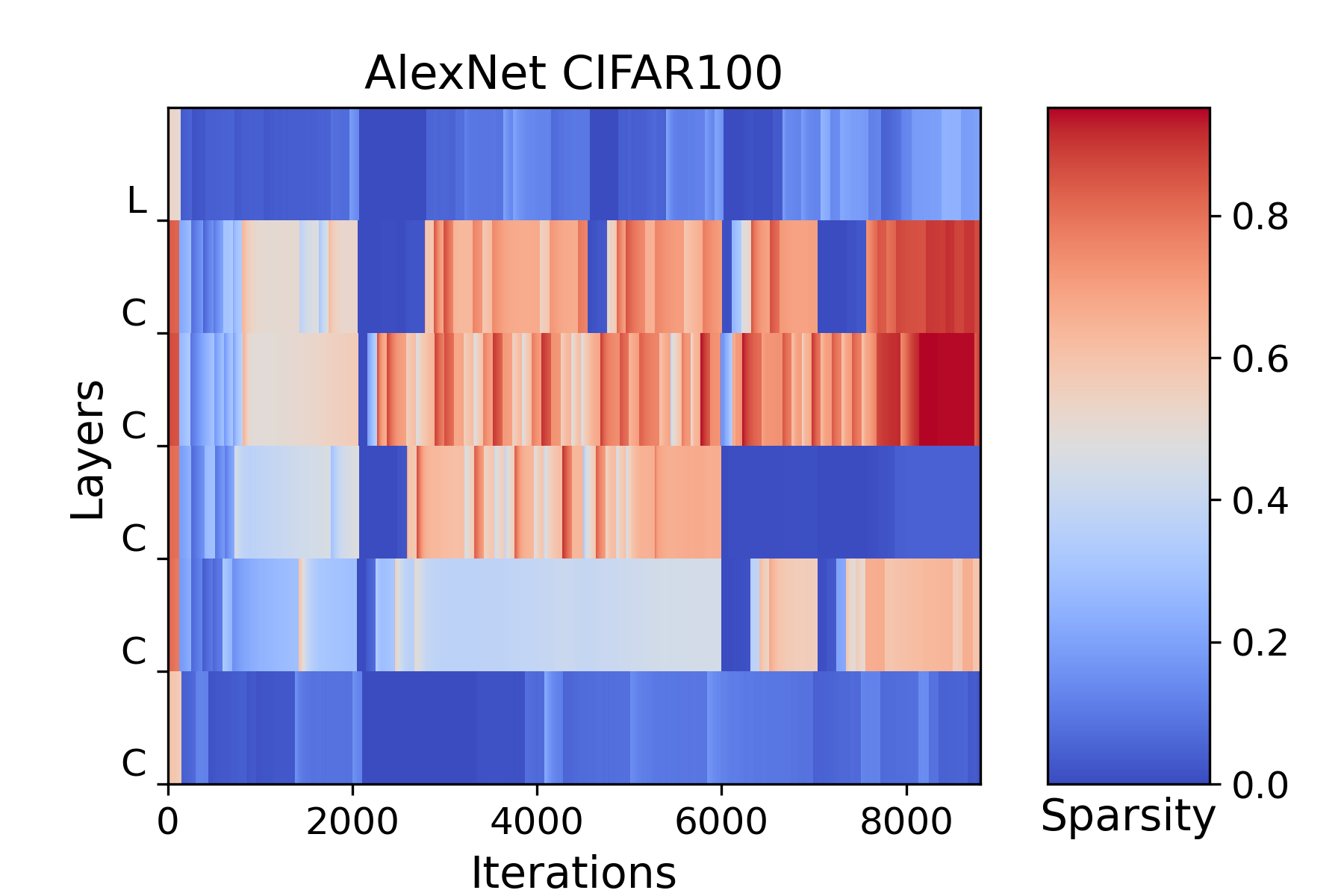}
\caption{\label{fig:eval_alexnet_c100_sp}\textit{Sparsity of ASGD optimized AlexNet on CIFAR100, 100 epochs}}
\end{figure}
Tabs. \ref{tab:overview3} and \ref{tab:overview4} show AdaPTs estimated speedups vs. our float32 baseline (100 epochs, same number of gradient accumulation steps and batch size as AdaPT trained models) and MuPPETs float32 baseline. AdaPT achieves speedups comparable with SOTA solutions on our own baseline and outperforms MuPPET on MuPPETs baseline in every scenario. Unfortunately, MuPPETs code base could not be executed so we were unable to apply MuPPET to the more efficient AdaPT baseline and were limited to comparing against results provided by the original authors. We used our performance model for simulating MuPPETs performance based on the precision switches stated in the MuPPET paper because MuPPETs authors did not publish their performance model. An overview over the reduction of computational costs through ASGD relative to a float32 baseline is provided by fig .\ref{fig:asgd_costs} as well
\begin{figure}[!h]
\includegraphics[width=0.5\textwidth]{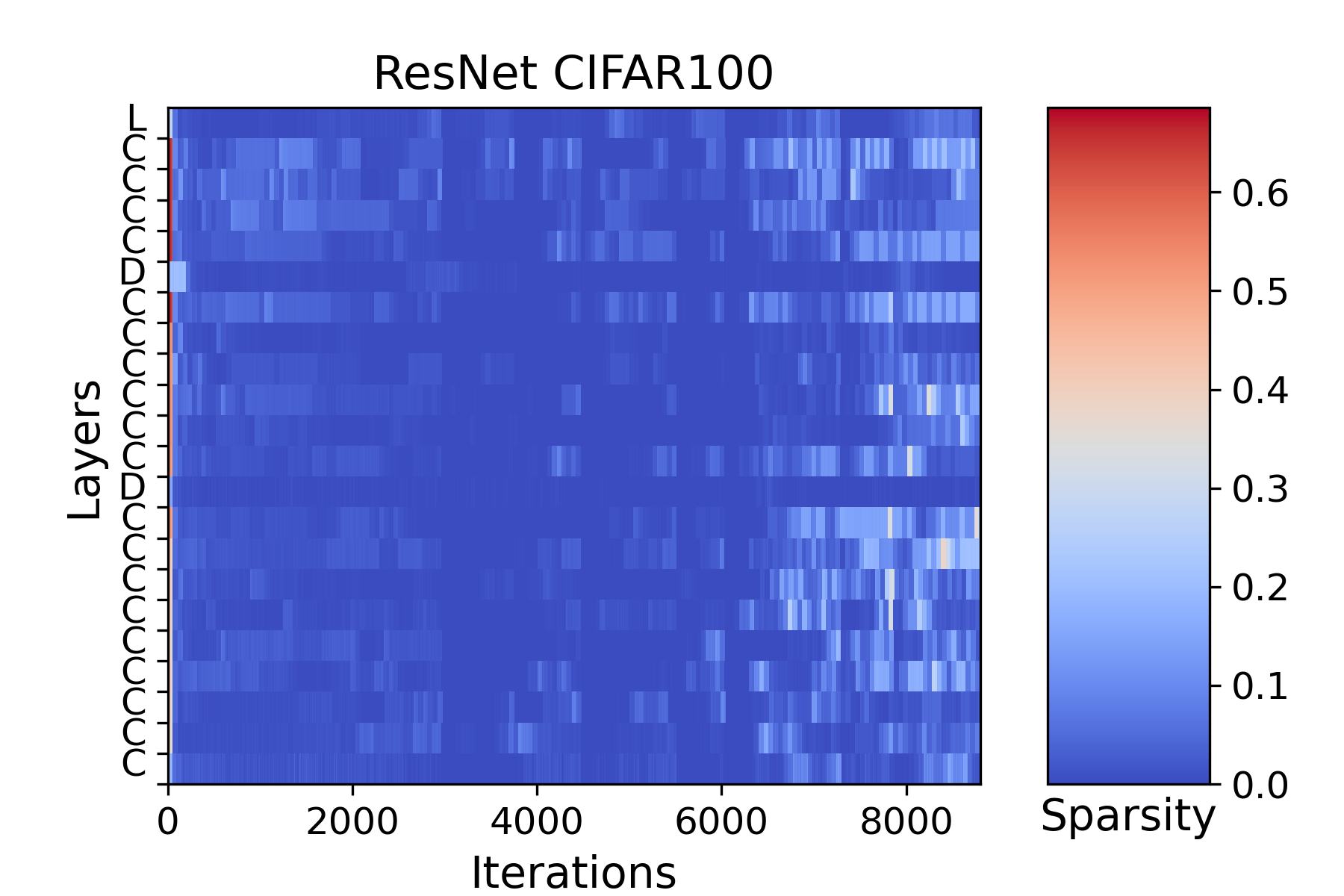}
\caption{\label{fig:eval_resnet_c100_sp}\textit{Sparsity of ASGD optimized ResNet on CIFAR100, 100 epochs}}
\end{figure}
Fig. \ref{fig:eval_alexnet_c100_sp} and fig. \ref{fig:eval_resnet_c100_sp} illustrate the induction of sparsity during AdaPT training. Interestingly in some cases, we observe an increasing degree of sparsity as AdaPT training progresses, with some layers reaching $80\%$ sparsity or more at the end of training as is the case with AlexNet trained on CIFAR100, while less sparsity could be induced in fewer layers training ResNet on CIFAR10/100. As can be seen in tab. \ref{tab:sparsity1}, AdaPT induces most sparsity in AlexNet, with a $45\%$ final model sparsity and an average intra-training sparsity of $34\%$ for training on CIFAR100. In the residual architecture ResNet, we observe that AdaPT introduces $7\%$ sparsity in the final model and $3\%$ intra-training sparsity. As fig. \ref{fig:eval_resnet_c100_sp} illustrates increasing sparsity towards the end training, we conjecture that for ResNet-like architectures, the sparsity inducing effect could be more pronounced if training is conducted for a higher number of epochs.
\begin{table}
\centering
\begin{tabular}{|l|ll|}
\hline
\textbf{Sparsity} & & \\
\textbf{} & \textbf{Final Model} & \textbf{Average}\\ \hline
\textbf{AlexNet}\textsubscript{CIFAR10} & 0.26 & 0.27 \\ \hline
\textbf{ResNet}\textsubscript{CIFAR10}  & 0.07 & 0.04 \\ \hline
\textbf{AlexNet}\textsubscript{CIFAR100} & 0.44 & 0.35 \\  \hline
\textbf{ResNet}\textsubscript{CIFAR100} & 0.07 & 0.03\\ \hline
\end{tabular}
\caption{\label{tab:sparsity1} Final model sparsity and average intra-training sparsity for AdaPT Training}
\end{table}
\begin{figure}[!h]
\includegraphics[width=0.5\textwidth]{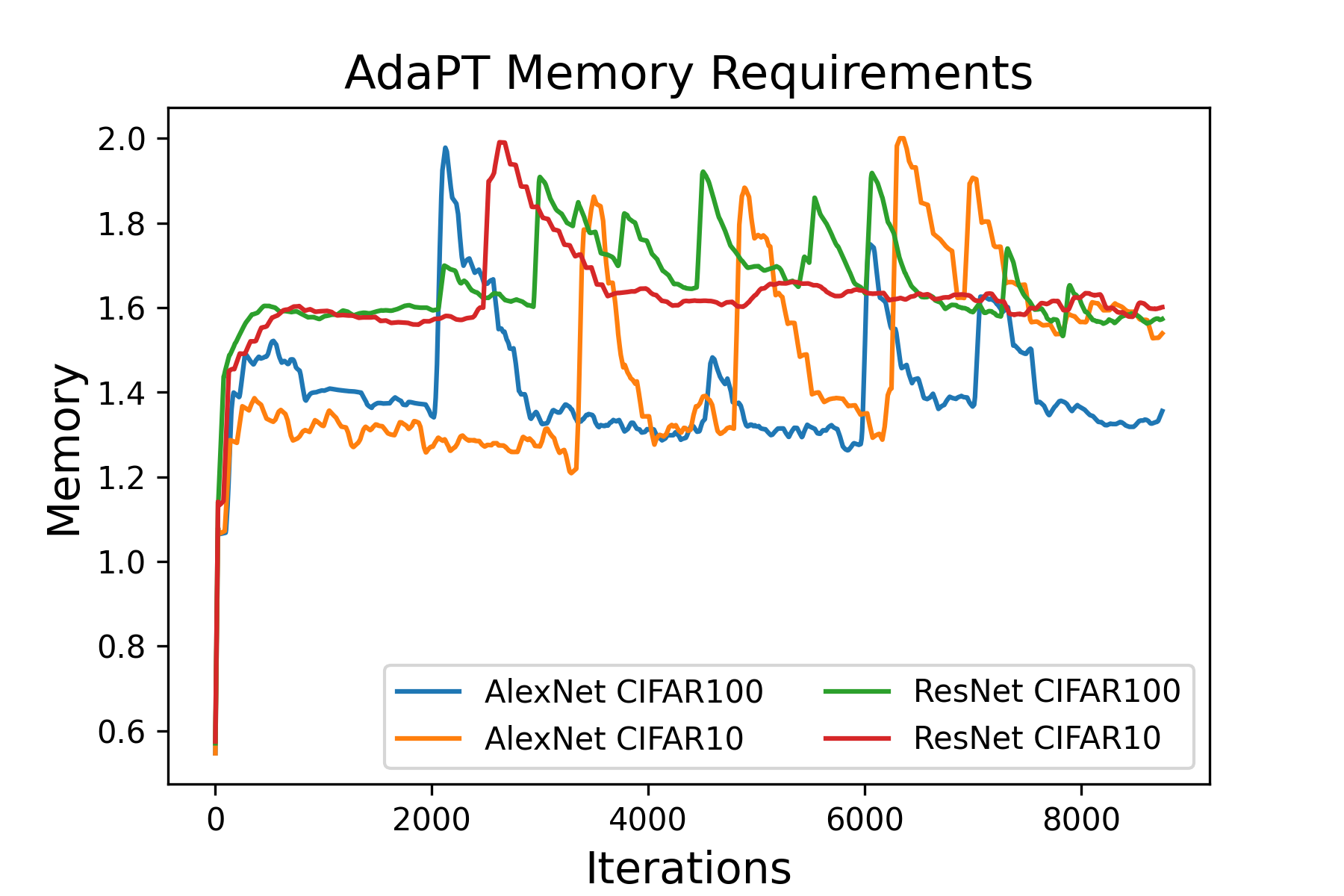}
\caption{\label{fig:asgd_mem}\textit{ASGD
Memory Consumption relative to float32 SGD}}
\end{figure}

A side effect illustrated by fig. \ref{fig:asgd_mem} and tab. \ref{tab:overview3} and tab. \ref{tab:overview2} of AdaPT for reducing the computational cost of network training (fig. \ref{fig:asgd_costs}) is it's increased intra-training memory consumption that is caused by AdaPT maintaining a complete float32 mastercopy of the models weights tensors used for backwards pass updates. However this effect is only present during training as the master-copies are discarded once the model is deployed and used for inference.
\subsubsection{Inference}
Given AdaPT trained networks are fully quantized and sparsified, they have advantage that extends beyond the training phase into the inference phase. Tab. \ref{tab:inf1} shows that inference speed ups for AdaPT trained networks range from $1.63$ to $3.56$. The speed ups achieved during the inference phase are even higher than those achieved during training because during inference, no expensive backwards pass has to be conducted and no overhead induced by AdaPT is occurring. This is a non-trivial advantage over MuPPET which does not provide any post-training advantages in terms of speed up or memory consumption at all due to the resulting network being float32.
\begin{table}
\centering
\begin{tabular}{|l|ll|}
\hline
\textbf{Inference} & & \\
\textbf{} & \textbf{SZ} & \textbf{SU}\\ \hline
\textbf{AlexNet}\textsubscript{CIFAR10} & 0.54 & 3.56 \\ \hline
\textbf{ResNet}\textsubscript{CIFAR10}  & 0.60 & 1.63 \\ \hline
\textbf{AlexNet}\textsubscript{CIFAR100} & 0.36 & 2.60 \\  \hline
\textbf{ResNet}\textsubscript{CIFAR100} & 0.57 & 1.52\\ \hline
\end{tabular}
\caption{\label{tab:inf1} Inference with AdaPT trained models}
\end{table}

\section{Conclusion}
\label{sec:conclusion}
With AdaPT we introduce a novel quantized DNN sparsifying training algorithm that can be combined with iterative gradient based training schemes. AdaPT exploits fast fixed-point operations for forward passes, while executing backward passes in highly precise float32. By carefully balancing the need to minimize the fixed-point precision for maximum speedup and minimal memory requirements, while at the same time keeping precision high enough for the network to keep learning, our approach produces top-1 accuracies comparable or better than SOTA float32 techniques or other quantized training algorithms. Furthermore AdaPT has the intrinsic methodical advantage of not only training the network in a quantized fashion, but also the trained network itself is quantized and sparsified s.t. it can be deployed to fast ASIC hardware at reduced memory cost and executed with reduced computational compared to a full precision model. Additionally we contribute a performance model for fixed point quantized training.
\section{Future Work}
\label{sec:futurework}
AdaPT as presented employs a fixed-point representation to enable fine granular precision switches. However, fixed-point hardware is not as common as floating-point hardware, so we plan to extend the concept to floating point quantization s.t. AdaPT becomes compatible with float16/float32 consumer hardware. Further, we intend to explore whether AdaPT can be used to generate very low bit-width (binarization/ternarization) networks through gradually reducing the quantization search space during training. Additionally, we conjecture the heuristics used by AdaPT can be used for intra-training DNN pruning as well, which will be subject to future research. We plan ablation testing to reduce the complexity of AdaPT. Another interesting application of AdaPT which we intend to explore is it's usefulness in other fields of research such as Drug Discovery. 

\section{Acknowledgements}
We thank Prof. Torsten Möller for sponsoring the DGX-1 GPU Cluster used in our experiments and we thank Prof. Sebastian Tschiatschek for feedback and useful discussions. Both were staff of the University of Vienna's Faculty of Computer Science at the time of writing.


\bibliographystyle{ieeetr}
\bibliography{marvin}

\end{document}